\documentclass{article}

\usepackage[final, nonatbib]{nips_2018}

\usepackage[utf8]{inputenc} 
\usepackage[T1]{fontenc}    
\usepackage{hyperref}       
\usepackage{url}            
\usepackage{booktabs}       
\usepackage{amsfonts}       
\usepackage{nicefrac}       
\usepackage{microtype}      
\usepackage{graphicx}

\title{Investigating performance of neural networks and gradient boosting models approximating microscopic traffic simulations in traffic optimization tasks}

\author{
  Paweł Gora\thanks{http://www.mimuw.edu.pl/\textasciitilde pawelg}\\
  TensorCell\\
  Faculty of Mathematics, Informatics and Mechanics\\
  University of Warsaw\\
  \texttt{p.gora@mimuw.edu.pl}\\
  \And
  Maciej Brzeski\\
  TensorCell\\
  Faculty of Mathematics and Computer Science\\
  Jagiellonian University\\
  \texttt{maciej.brzeski@doctoral.uj.edu.pl}\\
  \And
  Marcin Możejko\\
  TensorCell\\
  \texttt{mmozejko1988@gmail.com}
  \And
  Arkadiusz Klemenko\\
  TensorCell\\
  \texttt{arek.klemenko@gmail.com}
  \And
  Adrian Kochański\\
  TensorCell\\
  \texttt{ad.kochanski@gmail.com}
}

\begin{document}

\maketitle

\begin{abstract}
 We analyze the accuracy of traffic simulations metamodels based on neural networks and gradient boosting models (LightGBM), applied to traffic optimization as fitness functions of genetic algorithms. Our metamodels approximate outcomes of traffic simulations (the total time of waiting on a red signal) taking as an input different traffic signal settings, in order to efficiently find (sub)optimal settings. Their accuracy was proven to be very good on randomly selected test sets, but it turned out that the accuracy may drop in case of settings expected (according to genetic algorithms) to be close to local optima, which makes the traffic optimization process more difficult. In this work, we investigate $16$ different metamodels and $20$ settings of genetic algorithms, in order to understand what are the reasons of this phenomenon, what is its scale, how it can be mitigated and what can be potentially done to design better real-time traffic optimization methods.
\end{abstract}

\section{Introduction}
Vehicular traffic in cities is a complex process, it involves many components (e.g., vehicles, pedestrians, traffic lights) which are in a constant interaction. Thus, it is not easy to manage it optimally.

In this paper, we analyze an interesting phenomenon which we recently discovered in one of existing traffic optimization approaches, in which neural networks were used as metamodels approximating outcomes of time-consuming traffic simulations in order to speed up the process of finding optimal traffic signal settings \cite{gora2016, gora2017}. In that approach, neural networks are applied to replace simulations as fitness functions evaluating signal settings in a genetic algorithm (GA), which aims to efficiently find settings minimizing the total time of waiting on a red signal for a given configuration of cars (it is a so called, Traffic Signal Setting problem \cite{gora2015}). It turns out that the accuracy of traffic metamodels based on neural networks, which, in general, is very good in terms of performance on any randomly selected test set, decreases in case of settings found after many iterations of a GA, which makes the optimization process much more difficult. Luckily, genetic algorithms are still able to find relatively good settings (which still make the approach suitable for a real-world traffic management), but their general performance is disrupted because of the aforementioned phenomenon. 

In our investigation, we analyzed 8 metamodels based on neural networks and 8 based on LightGBM. First, we trained them on a training set consisting of $85336$ evaluations of different traffic signal settings, then, we used them as fitness functions in genetic algorithms aiming to find optimal settings. In total, we investigated $20$ configurations of parameters of genetic algorithms to examine how the phenomenon depend on values of different parameters.

The general conclusion is that the phenomenon is universal, although, its scale and properties depend on specific models (e.g., neural networks, LightGBM) and values of hyperparameters (e.g., type of an activation function). We present the detailed analysis of this phenomenon, its anticipated justification and expected approaches to mitigate its disruptive effect on traffic optimization. 

The rest of the paper is organized as follows: Section \ref{sec:relatedWorks} contains a review of related research works, Section \ref{sec:experiments} describes our experiments (settings, results and conclusions), Section \ref{sec:conclusions} concludes the paper.

\section{Related work}\label{sec:relatedWorks}
Black-box optimization (in which a black-box, such as simulator, is queried to obtain values for a specific input) is a well-known problem in machine learning \cite{amaran}. Many approaches have been used, but methods based on Bayesian optimization have become popular in recent years \cite{bayes}, especially when the query is expensive. It is a sequential approach, which tries to find the best point to receive true value in every step. In last years, more advanced methods using Reinforcement Learning \cite{RL} and Recurrent Neural Networks \cite{RNN} appeared - they try to learn how to choose such points. However these methods require incessantly querying, and it is hard to parallelize.

The presented research is inspired by experiments in \cite{gora2016} and \cite{gora2017}, in which neural networks were trained to approximate outcomes of traffic simulations and applied to the traffic optimization task (Traffic Signal Setting problem). That approach gave very good results, so we decided to investigate it in details and analyze its performance close to local minima found by genetic algorithms.

\section{Experiments}\label{sec:experiments}

\subsection{Setup of experiments}
For the purpose of our experiments, we trained 8 metamodels based on neural networks and 8 based on LightGBM. Below, we describe our dataset, metamodels and the genetic algorithm settings.

\subsubsection{Dataset}
The dataset used for training metamodels was generated using the Traffic Simulation Framework (TSF) \cite{gora2009}, similarly as in \cite{gora2016, gora2017} (TSF is able to run microscopic traffic simulations on realistic maps, consistent with the OpenStreetMap format, using a traffic model based on cellular automata). First, we selected an area in Warsaw consisting of 21 crossroads with traffic signals (Stara Ochota district). Then, we ran $105336$ simulations with randomly selected traffic signal settings as an input. Each traffic signal setting can be represented as a vector of $21$ elements, each corresponding to a traffic signal offset (being an integer value from the set \{0,1,\ldots , 119\}) for one of $21$ selected crossroads. For each setting, TSF computed the total time of waiting on a red signal in the selected area. For the purpose of experiments (i.e., training metamodels), we divided the set into a training set (consisting of the first $85336$ elements) and a test set (the remaining $20000$ elements). 

We made the dataset publicly available to scientific community to enable a further and more comprehensive research on this topic, the file with the whole dataset is available on the website \cite{dataset1} - each line consists of 21 offsets and the total time of waiting on a red signal computed using TSF for a given setting. It is also possible to get access to a microservice evaluating new settings using TSF, which was used to generate the training set (in that case, we ask for a direct contact with the authors of this paper).

\subsubsection{Metamodels based on neural networks}

Neural networks are powerful and highly adaptive machine learning algorithms. They are proven to be universal approximators \cite{cybenko} and achieve state-of-the-art results on many tasks despite significant overparametrization \cite{memoization}. In our experiments, we tested multiple fully-connected network architectures with and without residual connections \cite{resnet}, regularized using $l2$-weight regularization and Batch Normalization layers \cite{batchnorm}. Our models were trained using RMSProp optimizer \cite{gradientmethods}, for $200$ epochs with the batch size of $128$ and a mean squared error as the loss function. In order to normalize the input data and capture its periodic properties we transformed each coordinate \textit{x} of the input data to a pair $\left(\cos \frac{2\pi\textit{x}}{120}, \sin\frac{2\pi\textit{x}}{120}\right)$, what changed the input shape from $(21,)$ to $(42,)$. We trained 216 models with different values of regularization rates, number of units and layers, activations (ReLU \cite{relu} and tanh \cite{glorot}) and chose a set of 8 models (from the set of the best 30 models with the lowest relative error rates) which had the greatest variability in terms of model metaparameters. All models were trained using Keras \cite{keras} package with Tensorflow \cite{tensorflow} backend. The parameters of choosen models are presented in Table 1 in the Supplementary material 1 \cite{supp1}.

\subsubsection{Metamodels based on LightGBM}
Gradient boosting models belong to a family of machine learning algorithms, where the main idea assumes that weak predictive models become better with subsequent iterations, throughout learning from mistakes done by previous models. In our case, weak learners are represented by decision tree regressors. The most widely used implementations of this algorithm are XGBoost \cite{chen} and LightGBM \cite{ke}. The LightGBM algorithm is a gradient boosting framework, which uses tree based learning algorithms. In contrast to other tree based algorithms, which use level-wise tree growth, LightGBM grows trees vertically. It has been proven that LightGBM remarkably outperforms XGBoost in terms of memory consumption and computational speed \cite{ke}, which agrees with the outcomes of our experiments. These factors are vitally important in our research, as the execution time of a traffic metamodel should be as short as possible. On top of that, we managed to achieve slightly better accuracy with LightGBM, hence our choice is to base the further experiments upon this gradient boosting algorithm.

In order to find good values of hyperparameters, the Tree-structured Parzen Estimator was used, which is an efficient and robust hyperparameter optimization technique \cite{bergstra}. LightGBM has many parameters\footnote{LightGBM's parameters are explained on the website: \url{https://lightgbm.readthedocs.io/en/latest/Parameters.html}}, we initially selected a core set of hyperparameters and found their optimal values through training and validating the model on our training set, with 20\% split between train/test sets (see Table 2 in the Supplementary material 1 \cite{supp1}).
    
Our idea was to create $8$ different models and analyze how they perform comparing to real TSF's outcomes, trying also to find some patterns in their behaviour. We decided to keep optimized values of aforementioned hyperparameters and experiment with some other important parameters. Since LightGBM uses a leaf-wise growth algorithm, num\_leaves (number of leaves) controls its complexity. Regression\_l1 evaluates a model based on the mean absolute error metric, regression\_l2 and Poisson use root mean squared error and negative log-likelihood, respectively, as an evaluation metric. Apart from traditional Gradient Boosting Decision Tree (gbdt) method, we also included Dropouts meet Multiple Additive Regression Trees (dart) method in our experiments. Different values of num\_leaves, regressions and boosting techniques influence results strongly, so we decided to play with these parameters, train multiple models and select best $8$ models. Table 3 in the Supplementary material 1 (\cite{supp1}) presents the selection of chosen models along with relative errors made on our test set.

\subsubsection{Settings of genetic algorithms}
The investigated Traffic Signal Setting problem belongs to the family of nonconvex combinatorial optimization problems and for such tasks it is often difficult to find global optimum and even finding suboptimal solutions may be challenging, especially when obtaining value of the utility function evaluating traffic signal settings is also computationally expensive and that’s our case. Usually, the best approach is to use metaheuristics \cite{luke} - algorithms searching space of possible solutions, which make decisions which points to evaluate next based on past experience, e.g., genetic algorithms, simulated annealing, Bayesian optimization, tabu search, particle swarm optimization. We conducted experiments with these algorithms and realized that genetic algorithms outperforms other methods in our task. Therefore, we decided to focus in next experiments only on genetic algorithms.

In our use case, we represented traffic settings as 21-element vectors of offsets of traffic signals, being integer numbers from the set \{0,1,2,...,119\} (e.g. [4,4,48,82,41,57,35,54,54,14,30,31,31,31,92,111,13,13,77,110,4]) and we evaluated them using fitness functions representing computed (by TSF) or estimated (by neural networks or LightGBM models) total time of waiting on a red signal.

For conducting experiments using genetic algorithms, we developed a dedicated Python library. Thanks to that, we were able to test 504 configurations of hyperparameters of our algorithms (all configurations are described in the Supplementary Material 2 \cite{supp2}). For each configuration, we ran experiments for each of our $16$ models, $5$ times per each model (each run starts from a random initial population, so we wanted to test different initial settings), giving $2520$ runs for each model.

We had observed earlier that evaluating genotypes using TSF takes about $30$ seconds, which is too long, but in case of evaluations using metamodels (neural networks and LightGBM) the time of inference was much lower (less than 0.1 second), so we were able to conduct such compute-intensive experiments.

\subsection{Results of experiments}

\subsubsection{Initial experiments}

In the past experiments, genetic algorithms prove to be very efficient in finding suboptimal solutions for the Traffic Signal Setting problem (\cite{gora2016}, \cite{gora2017}), so it was no surprise that they were able to find good solutions in our case (at least, according to evaluation using metamodels). Usually, genetic algorithms stopped converging after about $50-100$ iterations (we had realized that earlier, so we decided to set $100$ as the number of iterations). The average value of the fitness function (computed using TSF) in the training set was about $48923$, but genetic algorithms were able to find points with a value at the level $32000-33000$ according to metamodels - as was later realized, the optimal values returned by genetic algorithms usually differed from TSF outcomes more than the average error of metamodels (about $1.7\%-2\%$), but settings found by genetic algorithms were still relatively good and much better than randomly selected settings.

Since we realized that the accuracy of approximations using metamodels is worse near local optima, we decided to examine this phenomenon. For that purpose, we had to compute true values of simulations for a large number of traffic signal settings. Each run of a genetic algorithm produced about $120$ (number of setting in a population) x $100$ (number of iterations) = $12000$ settings (in some cases the number was lower, because some settings were shared among different populations) and since we wanted to investigate universality of the phenomenon and test different values of hyperparameters of metamodels and genetic algorithms, we had to carefully select which settings should be evaluated using TSF (a single evaluation using TSF took about $30$ seconds on standard machines, which may give large time of computations in case of large number of evaluated settings). 

For each of $16$ models, from $2520$ genetic algorithm runs we selected best $10$ (according to the final best setting) and additional $10$ randomly from $90$ runs on positions $11-100$ (according to best found setting using a given model), so later we were investigating only $320$ GA runs and for each run we decided to evaluate using TSF $100$ trajectory settings (best settings from each one of $100$ iterations) and $100$ best settings found by a genetic algorithm run across all iterations.

\subsubsection{Distribution of errors}

\begin{figure}
\centering
\includegraphics[scale=0.25]{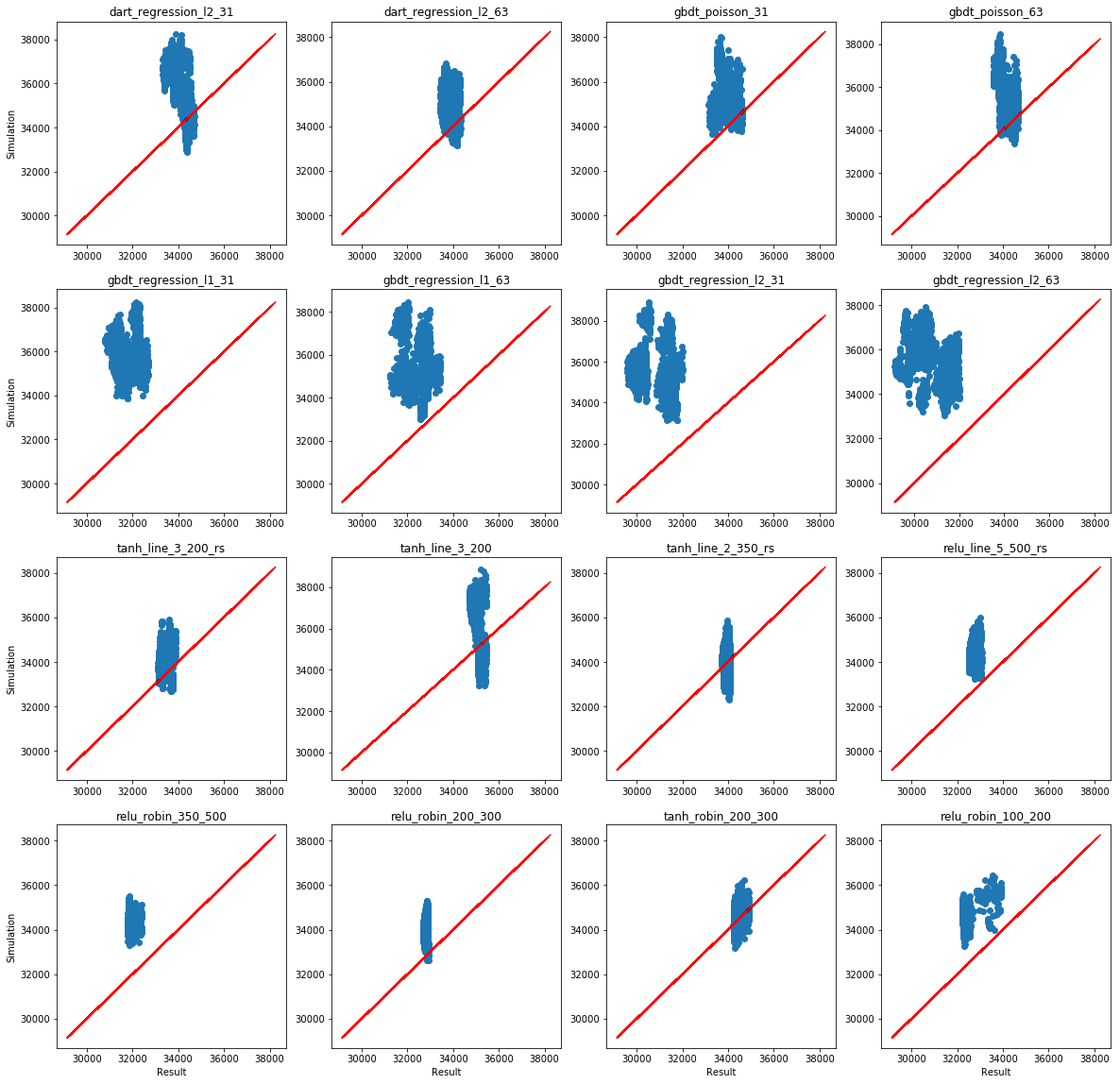}
\caption{Comparing model prediction and simulation for settings with the lowest predictions ($100$ best settings for $20$ genetic algorithm runs) for all $16$ investigated models.}
\label{fig:dist}
\end{figure}

For each model, we analyzed distributions of errors of approximations for best $100$ settings found in each of selected $20$ genetic algorithm runs (which gives $2000$ points, in total) - the summary is on the Figure \ref{fig:dist} presenting relations between approximations and true values returned by TSF.

One of interesting conclusions is that in most cases metamodels underestimate results of simulations and usually the error of approximation is larger than the average error of a given model on a random test set ($1.7-2\%$) and it can reach $12.9\%$ (in $14$ out of $16$ models it is larger than $2.16\%$), the same happens to the maximal error which can reach $21.5\%$.

Another interesting observation is that in case of neural networks activation function may significantly change the distribution of an error. For ReLU, neural network almost always underestimate results of simulations, for tanh the error is more symmetric.

We also analyzed errors for settings from GA trajectories and found that in case of last settings (which are also expected to be best) errors of approximations increase (Supplementary material 3 \cite{supp3}).

\subsubsection{Trajectories}
\begin{figure}
\centering
\includegraphics[scale=0.15]{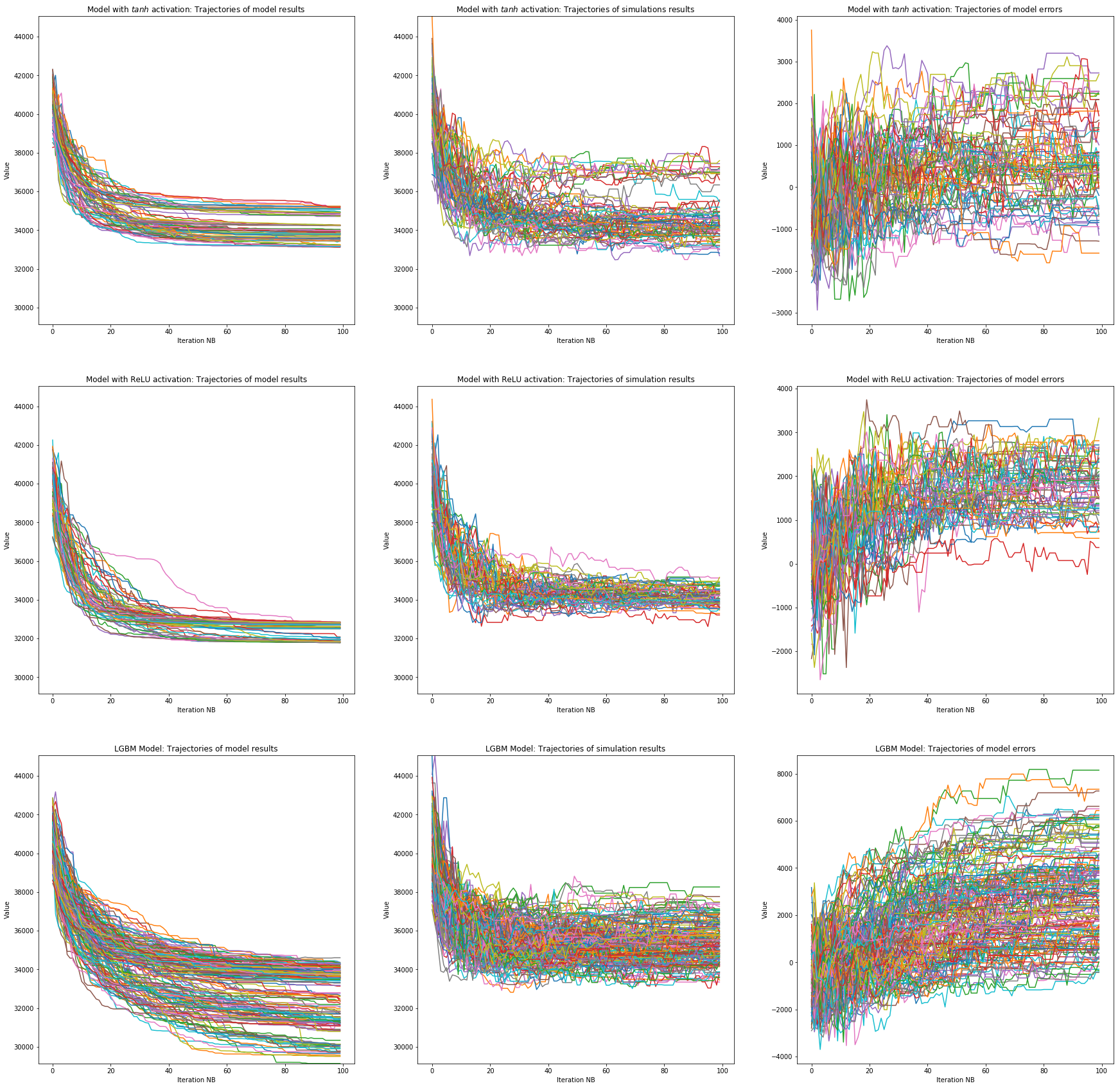}
\caption{Trajectories of GA runs, values of simulations and errors for neural networks with tanh (first row) and ReLU (second row) activation functions and for LightGBM (third row).}
\label{fig:trajectories}
\end{figure}

We also analyzed errors for traffic signal settings from trajectories (Figure \ref{fig:trajectories}). It confirms that, in general, error of approximations increases toward underestimating outcomes of simulations. Also, in case of LightGBM errors are becoming greater. A very important fact is that despite of increasing errors, genetic algorithms are still able to find much better settings than in the initial populations.

\subsubsection{Where do genetic algorithms converge to?}
\begin{figure}
\centering
\includegraphics[scale=0.15]{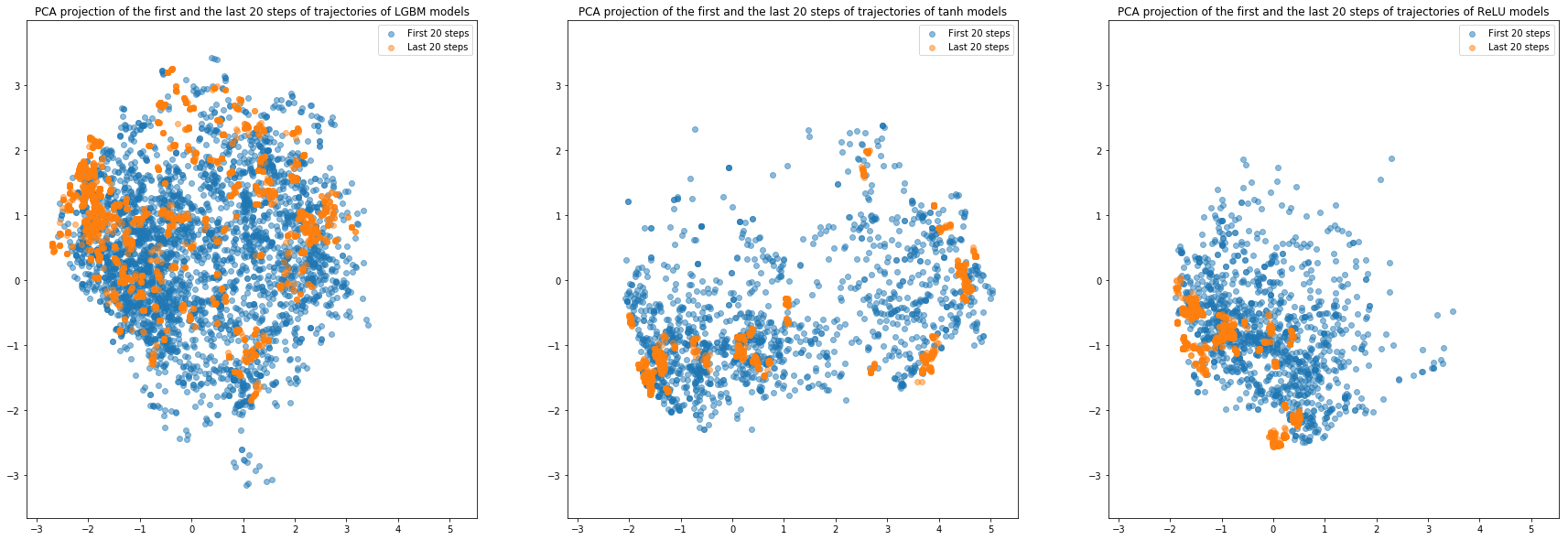}
\caption{Principal component analysis on best settings from different populations in GA}.
\label{fig:pca}
\end{figure}

We also ran a principal component analysis \cite{pca} on settings from trajectories and found out that, depending on the metamodel, genetic algorithms may converge to totally different points in the space of possible solutions (Figure \ref{fig:pca}).

\subsection{Discussion of results}
Based on our analysis, we conclude that for traffic signal settings found by genetic algorithms, errors of approximations increase and the phenomenon is universal - it can be observed in case of many different metamodels (neural networks, LightGBM), but its scale and properties may differ depending on some parameters, e.g., activation function in case of neural networks. We see $2$ potential reasons of this phenomenon: 
\begin{itemize}
    \item underrepresentation of local minima found by genetic algotihm in the training set
    \item the fact that genetic algorithms seek local minima and, therefore, they may also find points for which values of approximation are minimal and far from real simulation results
\end{itemize}

\section{Conclusions and future work}\label{sec:conclusions}
In the paper, we investigated performance of metamodels (based on neural networks and LightGBM) approximating outcomes of traffic simulations. Metamodels were used as fitness functions in the traffic optimization task and we found out that genetic algorithms searching for optimal settings of traffic signals, end up in areas of subspaces of possible settings, in which distribution of the error of approximation is much different than the standard distribution on a test set (errors are larger), which makes the traffic optimization task more difficult. We carried out detailed analysis of this phenomenon and it turns out that some hyperparameters of metamodels (e.g., activation function) have different impact on results. Moreover, the principal component analysis showed that genetic algorithms may converge to totally different areas of the searched space, depending on metamodels.

We also started analyzing how this phenomenon, which disrupts the traffic optimization task, can be mitigated. We see at least $2$ interesting research directions:
\begin{itemize}
    \item active learning / transfer learning approach - adding to the training set new settings found by a genetic algorithm (we investigate methods which may help make decisions when to run a time-consuming simulation in order to evaluate new points) and retraining metamodels
    \item ensemble learning approach - training several different models and taking as an output the average outcome of some of these models
\end{itemize}

We've already started working on these topics, but we also invite the scientific community to join this research (which is especially important from the traffic optimization perspective) and that's why we share a dataset which was used in our research \cite{dataset1}.

\section*{Acknowledgement}
The research was supported by Microsoft Research Azure Award and funding from the European Research Council (ERC) under the European Union's Horizon 2020 research and innovation programme under grant agreement No 677651. We would like to express our gratitude to our colleagues from the TensorCell team, who helped us on earlier stages of the presented research: Karol Kurach, Marek Bardoński, Magdalena Kukawska, Mateusz Susik and Dawid Kopczyk.

\end{document}